\documentclass[10pt,twocolumn,letterpaper]{article}
\usepackage{cfgs/iccv}              %

\definecolor{iccvblue}{rgb}{0.21,0.49,0.74}
\usepackage[pagebackref,breaklinks,colorlinks,allcolors=iccvblue]{hyperref}
\usepackage[capitalize]{cleveref}
\crefname{section}{Sec.}{Secs.}
\Crefname{section}{Section}{Sections}
\Crefname{table}{Table}{Tables}
\crefname{table}{Tab.}{Tabs.}
\def\paperID{xxxx} %
\def\confName{ICCV}
\def\confYear{2025}
\def\PaperTitle{
Generative Modeling of Shape-Dependent Self-Contact Human Poses
}
\def\DBname{Goliath-SC\xspace} %
\def\Mname{PAPoseDiff\xspace}
\def\DBsubj{130\xspace}
\def\DBpose{383K\xspace}

\def\DBtrain{313K\xspace}
\def\DBevalSubj{9.7K\xspace}

\author{
Takehiko Ohkawa\affmark[1,2*],
Jihyun Lee\affmark[1,3*],
Shunsuke Saito\affmark[1], 
Jason Saragih\affmark[1],
Fabian Prada\affmark[1],\\
Yichen Xu\affmark[1],
Shoou-I Yu\affmark[1],
Ryosuke Furuta\affmark[2], 
Yoichi Sato\affmark[2],
and Takaaki Shiratori\affmark[1]\\
\vspace{-2mm}
\newline\\
\affaddr{
\affmark[1]Codec Avatars Lab, Meta\hspace{4mm}
\affmark[2]The University of Tokyo\hspace{4mm}
\affmark[3]KAIST
}\\
\normalsize{
Project page: \url{https://tkhkaeio.github.io/projects/25-scgen}
}
}

\usepackage{graphicx}
\usepackage{amsmath}
\usepackage{amssymb}
\usepackage{booktabs}
\usepackage{bbold}
\usepackage{multirow}
\usepackage{bm}
\usepackage{arydshln}
\usepackage{xspace}
\usepackage{url}
\usepackage{comment}
\usepackage{mathabx}
\usepackage{wasysym}
\usepackage{capt-of}
\usepackage{adjustbox}
\usepackage[dvipsnames]{xcolor}
\usepackage{times}
\usepackage{epsfig}
\usepackage{amssymb}
\usepackage{cuted}

\newcommand*{\affaddr}[1]{#1} %
\newcommand*{\affmark}[1][*]{\textsuperscript{#1}}

\definecolor{GreenColor}{rgb}{0.137,0.573,0.565}
\definecolor{OrangeColor}{rgb}{0.914,0.541,0.0.141}
\definecolor{PurpleColor}{rgb}{0.5,0,0.7}
\definecolor{BlueColor}{rgb}{0,0.725,0.949}
\definecolor{PinkColor}{rgb}{0.9843,0.19215,0.6}

\newcommand\blfootnote[1]{%
  \begingroup
  \renewcommand\thefootnote{}\footnote{#1}%
  \addtocounter{footnote}{-1}%
  \endgroup
}

\makeatletter
\newcommand{\figcaption}[1]{\def\@captype{figure}\caption{#1}}
\newcommand{\tblcaption}[1]{\def\@captype{table}\caption{#1}}
\newcommand{\customparagraph}[1]{\par{\noindent\textbf{#1:}}}
\newcommand{\textcite}[1]{``\textit{#1}''}

\makeatother

\def\tbf{\textbf}

\def\chi{Proceedings of the SIGCHI Conference on Human Factors in Computing Systems (CHI)}

\makeatletter
\DeclareRobustCommand\onedot{\futurelet\@let@token\@onedot}
\def\@onedot{\ifx\@let@token.\else.\null\fi\xspace}
\def\eg{\emph{e.g}\onedot} 
\def\ie{\emph{i.e}\onedot} 
 
 \def\vs{\emph{vs}\onedot}
 
\def\etal{\emph{et al}\onedot}

\makeatother

\newcommand{\train}{\textit{train}\xspace}
\newcommand{\eval}{\textit{eval}\xspace}

\usepackage{arydshln}
\usepackage{xcolor}
\usepackage{booktabs} %
\usepackage{algorithm}
\usepackage{algpseudocode}
\algnewcommand{\LeftComment}[1]{\Statex \(\triangleright\) #1}

\title{\PaperTitle}
\begin{document}
\maketitle
\begin{abstract}
    One can hardly model self-contact of human poses without considering underlying body shapes. 
    For example, the pose of rubbing a belly for a person with a low BMI leads to penetration of the hand into the belly for a person with a high BMI.
    Despite its relevance, existing self-contact datasets lack the variety of self-contact poses and precise body shapes, limiting conclusive analysis between self-contact poses and shapes.
    To address this, we begin by introducing the first extensive self-contact dataset with precise body shape registration, \tbf{\DBname}, consisting of \DBpose self-contact poses across \DBsubj subjects.
    Using this dataset, we propose generative modeling of self-contact prior conditioned by body shape parameters, based on a body-part-wise latent diffusion with self-attention.
    We further incorporate this prior into single-view human pose estimation while refining estimated poses to be in contact. 
    Our experiments suggest that shape conditioning is vital to the successful modeling of self-contact pose distribution, hence improving single-view pose estimation in self-contact.
    \blfootnote{*Work done during the internship.}
\end{abstract}

\section{Introduction}\label{sec:introduction}
\begin{figure}[t]
\centering
\includegraphics[width=0.9\hsize]{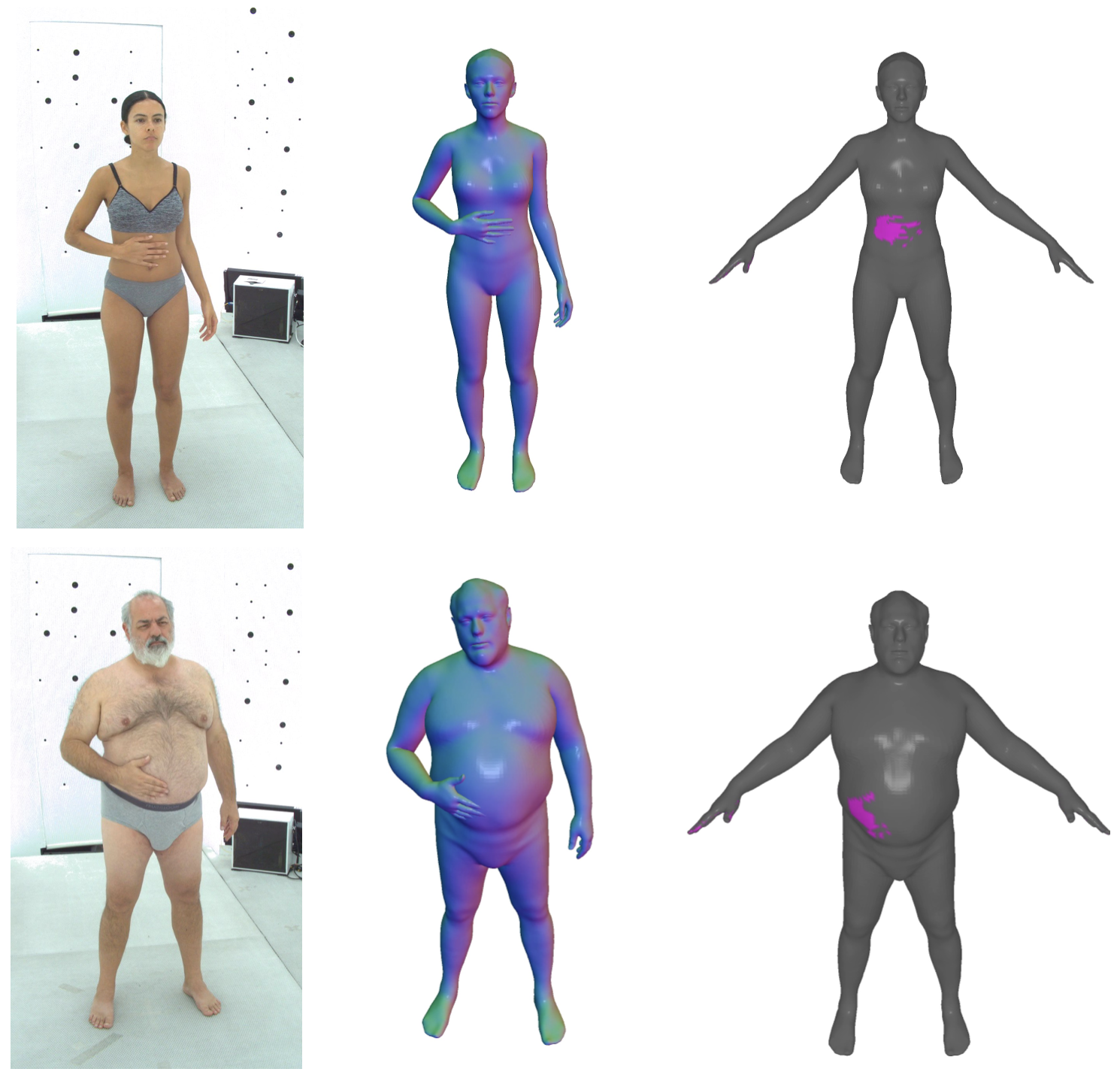}
\caption{
\tbf{Body shape dependency in self-contact poses}. 
We observe that self-contact poses (\eg, ``rubbing belly'') are influenced by the subject's body shape; for example, a person with a slimmer body (top) engages in different self-contact poses over one with a larger torso (bottom).
Indeed, the contact maps on the template mesh (right) differ noticeably.
Examples are sampled from the \tbf{\DBname} dataset we captured.
}
\label{fig:teaser}
\end{figure}

Human poses in our daily life often involve \textit{self-contact}, such as face touching, arm crossing, or hand placement, where body parts come into contact with the body surface.
These interactions with our own body are not only unconsciously displayed but also carry profound meaning across various disciplines, including psychology and social communication. 
Observing self-contact gestures, the areas touched can signal emotional states (\eg, anxiety or tension)~\cite{harrigan91,harrigan85,pease08}, and express linguistic symbols and contexts in sign language~\cite{uthus:nips23,yu:eccv24,chua:arxiv25}.
Notably, these self-contact poses are inherently constrained by the underlying body shapes.
As shown in \cref{fig:teaser}, two subjects performing ``rubbing belly'' gestures exhibit different poses and contacts due to variations in body shapes and proportions related to Body Mass Index (BMI).
Despite their significance, accurately modeling self-contact poses remains a considerable challenge~\cite{muller:cvpr21,fieraru:aaai21}; particularly, the dependency of human poses on body shapes is underexplored.

The challenge of self-contact modeling stems from the lack of datasets containing large self-contact poses with precise body shape registration.
Existing 3D body self-contact datasets, HumanSC3D~\cite{fieraru:aaai21} and MTP~\cite{muller:cvpr21}, contain small self-contact poses (1-4K poses) and suffer from inaccurate registration due to the lack of paired RGB images~\cite{muller:cvpr21}.
Other studies have highlighted specific part interactions, such as hand-hand~\cite{moon:eccv20} or hand-face contact~\cite{shimada:tog23}.
However, their scopes are limited to isolated body parts and fail to capture the holistic nature of self-contact, overlooking how the full-body pose and shape influence the contact.

Given the limitations of the existing datasets, we begin by offering the first extensive self-contact dataset with varying full-body poses and precise body shape registration, dubbed \tbf{\DBname}.
Our self-contact dataset contains the largest amount of self-contact poses, comprising \DBpose poses from \DBsubj subjects.
Additionally, it provides accurate full-body mesh registration based on 3D scans in a multi-camera dome (Goliath~\cite{martinez:nips24}), which are converted to SMPL-X~\cite{pavlakos:cvpr19} to access body shape parameters.

Using this dataset, we model the dependency of self-contact poses on body shapes via generative models. 
The generative modeling is designed to learn self-contact pose distribution for the given body shapes, independent of image input.
Pose-based generative training has an advantage over direct pose regression from images~\cite{fieraru:aaai21,muller:cvpr21,moon:eccv20,shimada:tog23} due to its generalizability to unknown environments and subjects.
Removing image input helps debias from the captured environments. 
Furthermore, it facilitates interpolation in the learned pose–shape space, enabling the modeling of plausible self-contact poses for novel body shapes or contact locations not explicitly seen during training.

In more detail, our approach involves a new insight of \textit{shape-dependency} in generative modeling with denoising diffusion.
Unlike joint distribution modeling between pose and shape~\cite{muller:cvpr24,lee:cvpr24} of 3D human models~\cite{loper:tog15,romero:tog17}, we explicitly model the shape-dependent manifold of self-contact poses using diffusion models~\cite{ho:nips20,song:iclr21}.
Specifically, we develop a latent diffusion model with self-attention, termed \tbf{\Mname}, which considers the relationship among highly interacting body parts (\eg, hands, body, and face).

Finally, we leverage the learned diffusion prior to refine 3D poses in self-contact.
Given the initial SMPL-X estimation, we refine the poses to have a smaller error to the 2D keypoint observation, while maintaining the plausibility in contact acquired by the former generative training. 
Our experiments demonstrate that our refinement with the shape-conditional diffusion prior surpasses a recent diffusion prior for human contact (BUDDI~\cite{muller:cvpr24}) and the state-of-the-art foundation model with direct regression (SMPLer-X~\cite{cai:nips23}) in the \DBname \eval set with unseen subjects.

Our contributions are summarized as follows:
\begin{itemize}
    \setlength{\parskip}{0pt}
    \setlength{\itemsep}{3pt}
    \item We introduce a new self-contact dataset \tbf{\DBname} with extensive poses and precise body shape registration.
    \item We propose generative learning of the shape-dependent manifold of self-contact poses, along with a latent diffusion with part-aware self-attention, \tbf{\Mname}.
    \item We propose an efficient single-view pose refinement that fits initial SMPL-X predictions to the observed 2D keypoints using the learned diffusion model. 
\end{itemize}

\begin{figure*}[t]
\centering
    \includegraphics[width=0.56\hsize]{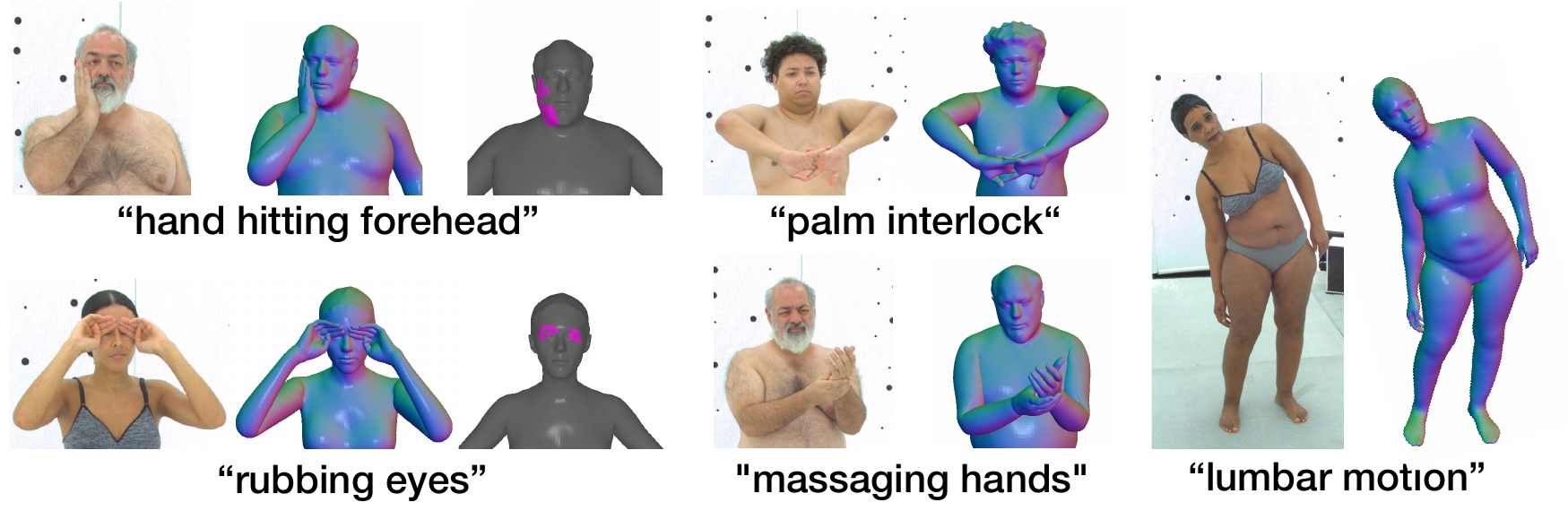}
\hfill
    \includegraphics[width=0.42\hsize]{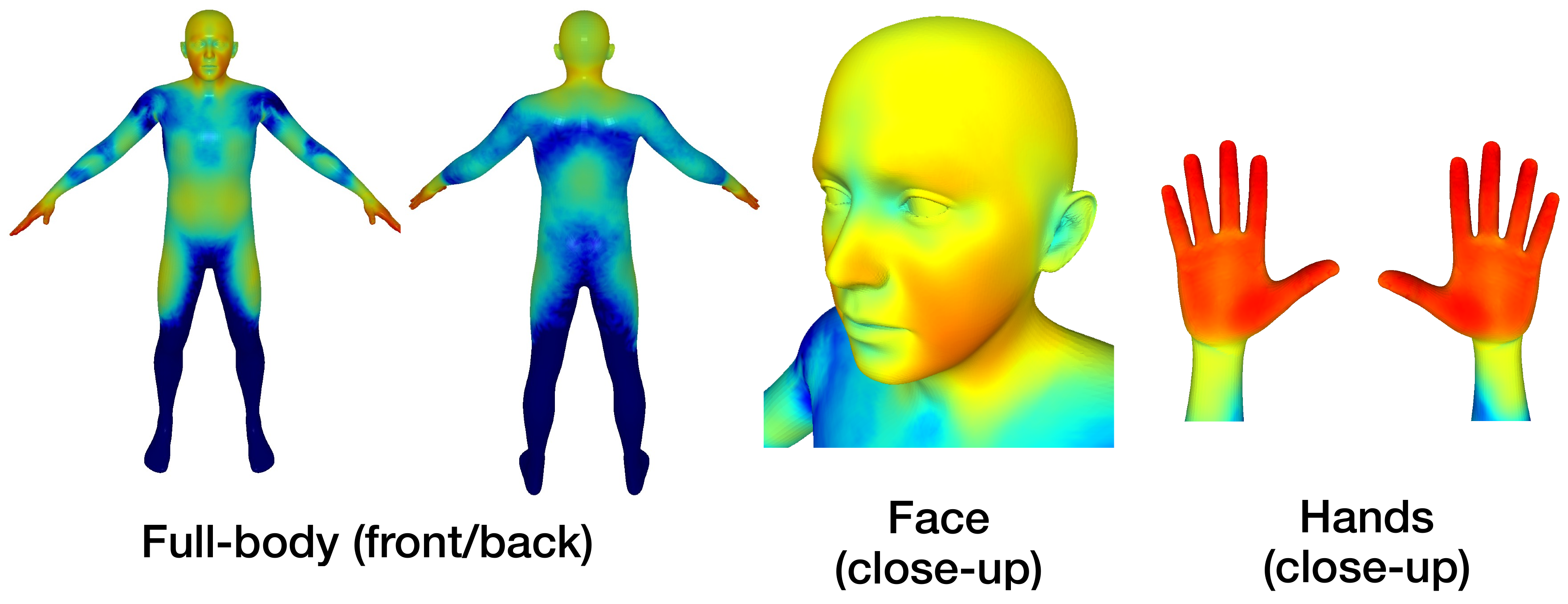}
\caption{\tbf{Examples of our \DBname dataset and contact heatmap.} We capture self-contact poses from \DBsubj subjects with scripted action instructions (\eg, ``hand hitting forehead''). 
Examples from the subjects of Goliath-4~\cite{martinez:nips24} are shown in the left figure.
We compute vertex-wise binary contact maps to find contact frames, and the averaged heatmap is shown in the right figure.
}
\label{fig:data}
\vspace{-4mm}
\end{figure*}

\section{Related Work}\label{sec:related-work}
\customparagraph{Self-contact datasets}
{
Human contact is taken into account in 3D human reconstruction~\cite{muller:cvpr21,yin:cvpr23,khirodkar:nips24}.
These studies include self-contact of a single person (\eg crossing arms)~\cite{muller:cvpr21,fieraru:aaai21}, multi-person interactions like hugging~\cite{fieraru:cvpr20,yin:cvpr23,khirodkar:nips24,muller:cvpr24,ho:vrst14}, or contact with external environments, such as scene~\cite{hassan:iccv19,huang:cvpr22,li:3dv22} and handheld objects~\cite{grady:cvpr21,chen3:cvpr23,fan:cvpr23,taheri:eccv20,ohkawa:cvpr23,liu:cvpr24,ohkawa:ijcv23,lin:iclr25}.
However, constructing self-contact datasets is particularly difficult due to higher self-occlusion. 
HumanSC3D~\cite{fieraru:aaai21} and MTP~\cite{muller:cvpr21} datasets contain a limited number of poses (1-4K poses) and inaccurate annotations due to the absence of paired RGB images with the captured poses~\cite{muller:cvpr21}.
Other studies focus on specific part interactions; InterHand2.6M captures hand-hand interactions~\cite{moon:eccv20}, while Decaf highlights hand-face contact~\cite{shimada:tog23}.
Despite allowing fine interaction analysis, capturing only isolated parts disregards the holistic perspective of self-contact, \ie, how the body influences hands and face in contact.
In contrast, our captured \DBname dataset provides extensive self-contact poses (\DBpose) with a dense camera setup, including full-body registration with precise shapes.
Our dataset also includes continuous pose variations, unlike frame-independent pose registration in MTP~\cite{muller:cvpr21}.
}

\customparagraph{Self-contact estimation}
{
Previous self-contact works follow a \textit{regressive} approach, aiming to estimate contact states from a single image.
Early attempts~\cite{fieraru:aaai21,muller:cvpr21} rely on the annotation of discrete 2D contact labels, representing which body parts are in contact, though the annotation process is labor-intensive and difficult to scale.
Fieraru~\etal formulate the tasks of segmenting in-contact parts and predicting interacting part pairs (contact signature)~\cite{fieraru:aaai21}.
Muller~\etal estimate human poses in self-contact~\cite{muller:cvpr21}, with two distinct training setups: (1) supervised training of a regressor on 3D GTs (\ie, MTP~\cite{muller:cvpr21}) and (2) additional optimization when 3D GTs are unavailable, instead relying on discrete contact labels (\ie, in-the-wild data like DSC~\cite{muller:cvpr21}).
Without using such contact labels, recent human foundation models (\eg, SMPLer-X~\cite{cai:nips23}) extend (1)'s approach to train a ViT network~\cite{dosovitskiy:iclr21} across various 3D human datasets, including self-contact scenarios (\ie, HumanSC3D~\cite{fieraru:aaai21} and MTP~\cite{muller:cvpr21}).
While this simple regression strategy generalizes across domains, it still struggles to capture the nuanced self-contact.
To address this, we investigate a \textit{generative} approach to refine the regressor's estimates, without relying on image input and manual annotations for 2D contact parts.
}

\customparagraph{Diffusion models}
Denoising diffusion~\cite{song:iclr21,ho:nips20} is becoming a popular choice for generative prior modeling due to its higher capability compared to handcrafted methods~\cite{akhter:cvpr15,pavlakos:cvpr19} or VAEs~\cite{kingma2:iclr14,pavlakos:cvpr19}.
Diffusion models are trained to iteratively denoise a Gaussian noise to sample from the learned data distribution~\cite{ho:nips20,song:iclr21}.
While they have been widely adopted, \eg, for motion synthesis~\cite{tevet:iclr23,xu:iccv23}, only a few works have modeled human contact using diffusion models. 
BUDDI~\cite{muller:cvpr24} and InterHandGen~\cite{lee:cvpr24} are concurrently proposed to model the contact between two bodies (either human bodies or hands) with the DDPM formulation~\cite{ho:nips20}.

While these previous methods learn the joint distribution of the pose and shape parameters of SMPL~\cite{loper:tog15} or MANO~\cite{romero:tog17}, our diffusion modeling relies on a new assumption that \emph{pose should depend on body shape}, thus generating poses conditioned on the given body shapes.
Furthermore, when adapting the diffusion prior to single-view pose estimation, our proposed refinement-based method does not require additional fine-tuning as in \cite{lee:cvpr24}.

\begin{table}[t]
\centering
\resizebox{1.0\linewidth}{!}{
\begin{tabular}{l|cccc}
\toprule
Dataset                          & \#SCPose   & \multicolumn{1}{l}{\#Subj.}                           & Params                                                                                           & Annot.                                                                                  \\\hline
HumanSC3D~\cite{fieraru:aaai21}  & 1.0K     & \begin{tabular}[c]{@{}c@{}}6\\ (3/3/0)\end{tabular}     & \begin{tabular}[c]{@{}c@{}}SMPL-X~\cite{pavlakos:cvpr19}\\/ GHUM~\cite{xu:cvpr20}\end{tabular} & 
Mocap                                                                                   \\
FlickrSC3D~\cite{fieraru:aaai21} & $<$3.9K  & -                                                     & SMPL-X                                                                                           & Pseudo-3D-GTs                                                                           \\
MTP~\cite{muller:cvpr21}         & 1.6K     & \begin{tabular}[c]{@{}c@{}}148\\(52/96/0)\end{tabular} & SMPL-X                                                                    & Pseudo-3D-GTs 
\\\hdashline
\vspace{-2mm}\\
\begin{tabular}[c]{@{}c@{}}\tbf{Goliath-SC} (Ours)\end{tabular} & 383K    & \multirow{2}{*}{\begin{tabular}[c]{@{}c@{}}130\\ (70/56/4)\end{tabular}}   & \multirow{2}{*}{SMPL-X}        & \multirow{2}{*}{\begin{tabular}[c]{@{}c@{}}MV RGB scan~\cite{martinez:nips24}\\ (cam: 220)\end{tabular}} \\   
\vspace{0mm}\\
\bottomrule
\end{tabular}
}
\caption{\textbf{Comparison of full-body self-contact datasets.}
We compare the number of self-contact poses, captured subjects, body parametrization, and annotation methods. The subject data include the gender ratio (female/male/non-binary).
}
\label{tab:data}
\end{table}

\section{Self-Contact Analysis and Dataset}\label{sec:data}
We introduce a new self-contact dataset with varying full-body poses and shapes, termed \tbf{\DBname}.
Our capture is based on a multi-camera dome setup of Goliath~\cite{martinez:nips24} with 3D full-body scans from 220 RGB cameras.
The scope of captured activities lies in natural self-contacts occurring in daily life like touching the face, body, hands, etc. 
\cref{tab:data} and \cref{fig:data} show data statistics and examples.

Our dataset has the following advantages.
(1) Our captures contain substantial self-contact data with \DBpose poses from \DBsubj subjects, exceeding the existing self-contact datasets~\cite{fieraru:aaai21,muller:cvpr21} by two orders of magnitude.
(2) Owing to numerous cameras with increased resolutions for the face and hands areas~\cite{martinez:nips24}, it provides high-quality mesh registration with fine details for hands and face.
This enables capturing fine self-contacts like ``rubbing eyes'' and ``massaging hands'', which is distinguished from the existing self-contact scenarios~\cite{fieraru:aaai21,muller:cvpr21,fieraru:cvpr20,yin:cvpr23}.
(3) Instead of collecting in-contact poses independently~\cite{muller:cvpr21}, we capture the sequence of natural self-contact poses at 30~Hz with scripted action instructions (\eg, ``rubbing belly''), leading to diverse and continuous self-contact poses.
(4) Unlike targeting specific body parts (\eg, hand-hand in InterHand2.6M~\cite{moon:eccv20}, hand-face in Decaf~\cite{shimada:tog23}), our dataset provides complete 3D full-body poses including hand-hand and hand-face interactions.
This enables holistic behavior modeling in which hands and face are constrained by the body's kinematics.
(5) To model the shape-dependent manifold, we convert the registered meshes to SMPL-X~\cite{pavlakos:cvpr19}, which gives the latent shape parameters.
Details are found in the supplement.

\customparagraph{Contact maps and data screening}
To comprehend self-contact patterns, we compute vertex-wise contact maps; see \cref{fig:data}.
We first create binary contact maps by discriminating if the hand vertices are close ($< 3$mm) to the rest of the body vertices, and then select contact frames with positive contact maps of the captured sequence.
This indicates that each sample represents a unique self-contact pose in which the hand is touching somewhere on the body.

We then calculate the contact heatmap from the binary maps, indicating contact likelihood as~\cite{taheri:eccv20}.
We observe that it includes various interactions across hands, face, neck, arms, and torso.
While the studies on hand-object grasping have a high contact likelihood in the finger areas~\cite{fan:eccv24,fan:cvpr23,taheri:eccv20,ohkawa:cvpr23}, our self-contact data include frequent interactions in the palm of hands as well.
This suggests that self-touching gestures are expressed by using hands widely from the palm to the fingertips.

\begin{figure*}[t]
    \centering
    \includegraphics[width=0.52\hsize]{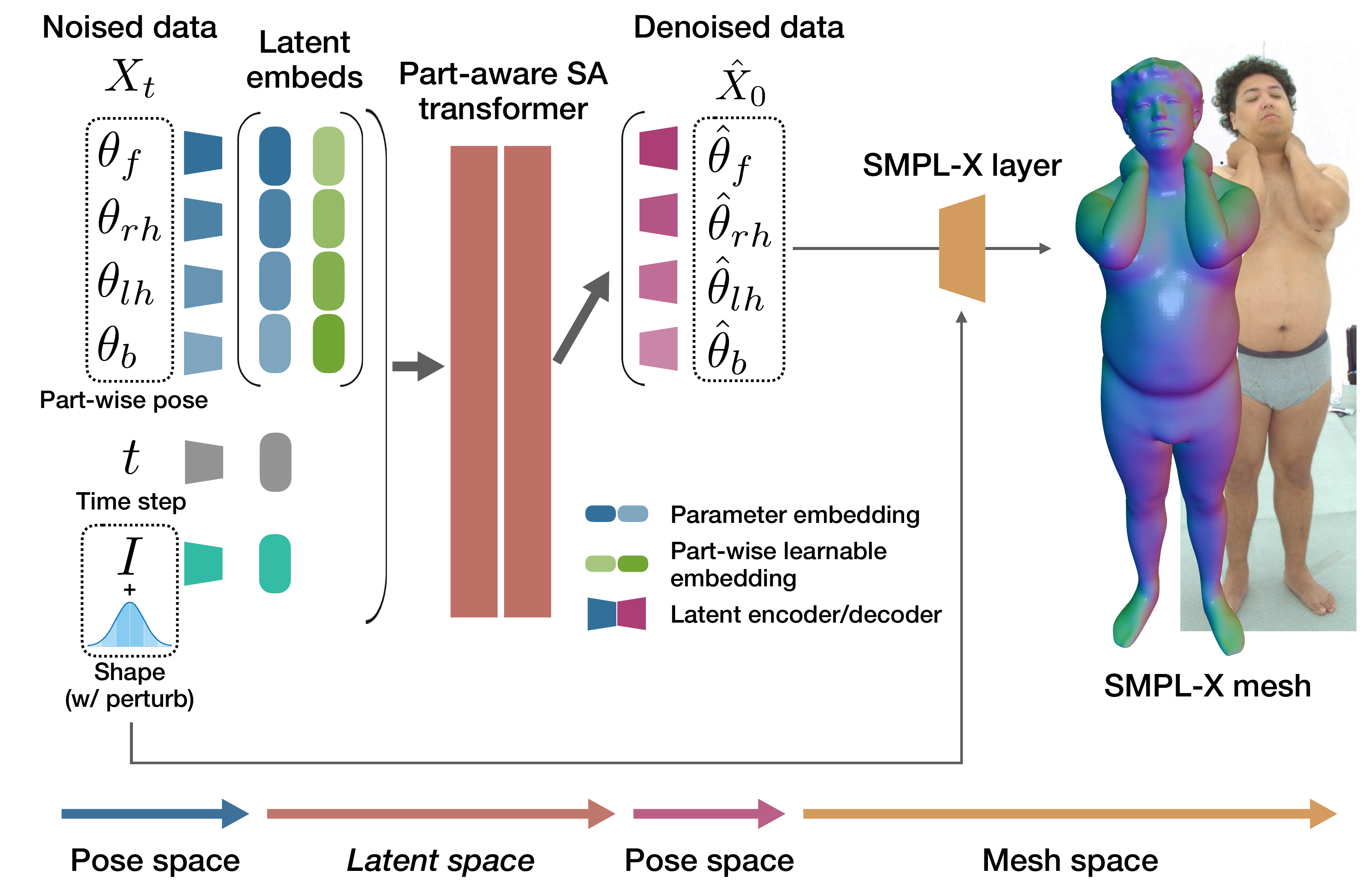}
    \hfill
    \includegraphics[width=0.46\hsize]{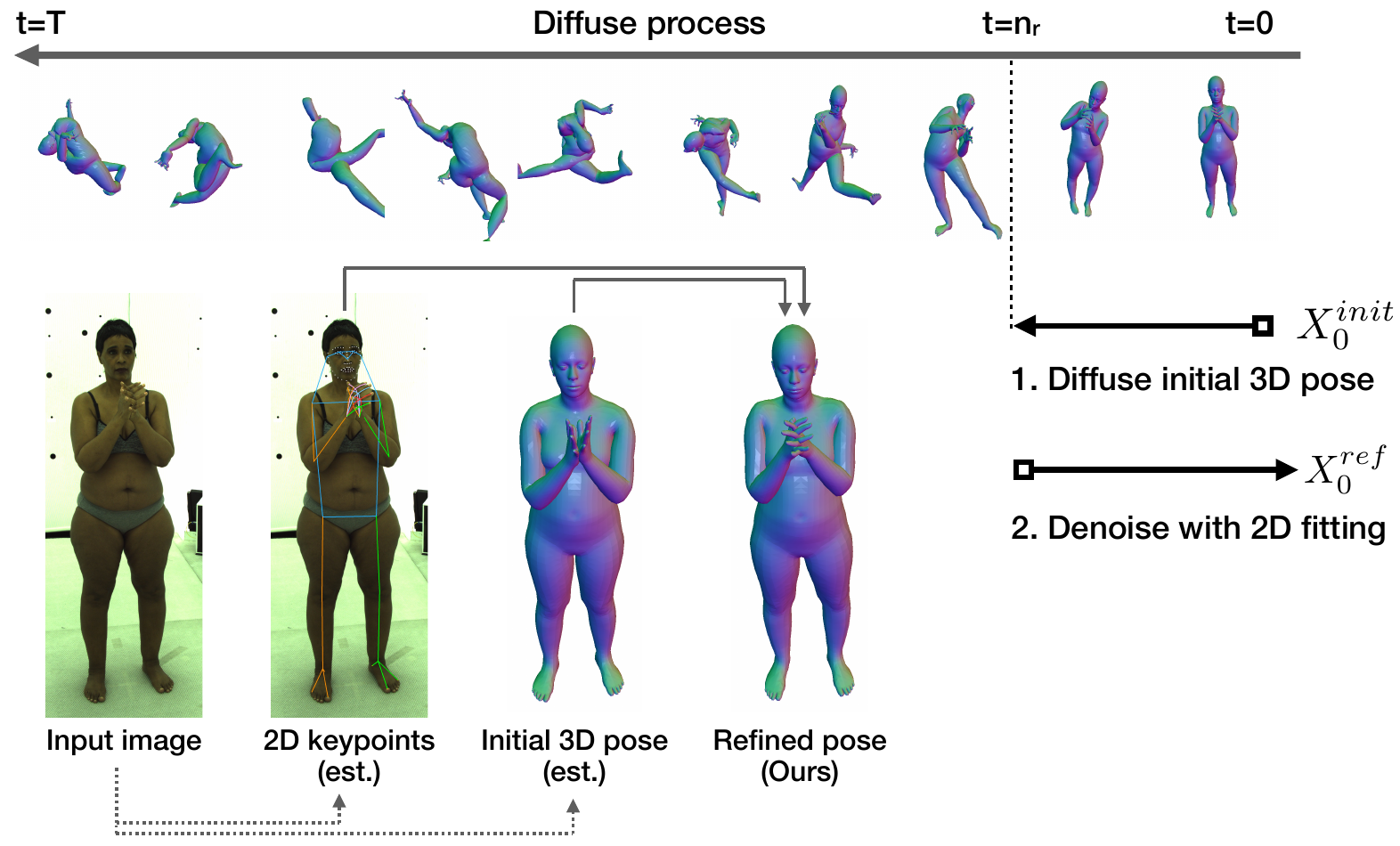}
\caption{\tbf{Shape-conditional denoising diffusion model for self-contact poses (left) and single-view refinement (right).}
Our proposed diffusion model (left), \tbf{\Mname}, follows latent diffusion with part-aware attention.
The model is trained to generate part-wise pose parameters conditioned on the shape information while considering their interactions with self-attention (SA).
We also add small perturbations for the shapes to generalize to unseen subjects. 
The training losses are described in \cref{sec:loss}.
Our refinement (right) is based on the observations of 2D keypoints and initial 3D pose estimation. We diffuse the initial 3D pose $\boldsymbol{X}^{init}_0$ and then denoise it to obtain a refined pose $\boldsymbol{X}^{ref}_0$ while fitting to the 2D observation.
}
\label{fig:method}
\vspace{-4mm}
\end{figure*}

\section{Method}\label{sec:proposal}
We present our proposed generative diffusion model for self-contact pose modeling.
Here, we aim to model the manifold of self-contact poses, particularly 
depending on the subject's body shape.
We detail our task and model setup in~\cref{sec:diff,sec:model} and training objectives in \cref{sec:loss}.
We then provide the inference process in~\cref{sec:infer}.
\cref{fig:method} shows the overview of our proposed diffusion model, dubbed \tbf{\Mname}, and our refinement scheme to obtain refined pose  $\boldsymbol{X}^{ref}_0$ given initial 3D pose estimate $\boldsymbol{X}^{init}_0$.

\subsection{Diffusion process}\label{sec:diff}
We follow the DDPM formulation~\cite{ho:nips20}, where the diffusion process consists of forward and reverse paths spanned with diffusion time steps $t \in [1,T]$.
The forward process ($1 \rightarrow T$) takes an input data $\boldsymbol{X}_0$ and gradually adds standard Gaussian noise $\epsilon_t$
to the data.
We denote the process of diffusing $\boldsymbol{X}_0$ at step $t$ as $\boldsymbol{X}_t=\mathrm{noise}\left(\boldsymbol{X}_0, t\right)$, formulated as
\begin{equation}
\mathrm{noise}\left(\boldsymbol{X}_0, t\right)=\sqrt{\bar{\alpha}_t} \boldsymbol{X}_0+\sqrt{\left(1-\bar{\alpha}_t\right)} \epsilon_t.
\end{equation}
The noisiness of $\boldsymbol{X}_t$ is controlled by noise variances $\beta_t$, \eg, given by a cosine scheduler~\cite{nichol:icml21}.
The coefficients that balance the noise and data terms are determined by $\alpha_t = 1 - \beta_t$ and $\bar{\alpha}_t=\Pi_{s=1}^t \alpha_s$.
$T$ is set to 1000.

The reverse process ($T \rightarrow 1$) denoises the data in every step and finally generates a clean sample ${\hat{\boldsymbol{X}}}_0$ with a learnable model $f$.
Following~\cite{song:iclr21,tevet:iclr23}, given the noised data $\boldsymbol{X}_t$ in step $t$, the model directly approximates the original data $\boldsymbol{X}_0$ as $\hat{\boldsymbol{X}}_0=f\left(\boldsymbol{X}_t, t\right)$.
For conditional generation, the model can take an additional conditional input $\boldsymbol{c}$ as $f\left(\boldsymbol{X}_t, t, \boldsymbol{c}\right)$.

\subsection{Shape-dependent pose modeling}\label{sec:model}
With the diffusion formulation, we construct data representations for poses and network architectures that enforce the shape-dependent constraint.
We explicitly model the interactions between different body parts including hands, body, and face, which distinguishes our approach from the previous studies that only use body parameters~\cite{muller:cvpr21,muller:cvpr24}.

\customparagraph{Data representation}
To learn self-contact poses, we use part-wise pose parameters as the target to denoise and shape parameters as conditional input, which are obtained by the differentiable SMPL-X model~\cite{pavlakos:cvpr19}.
The input pose data are constructed as $\boldsymbol{X}=\left[\boldsymbol{\theta}_{f}, \boldsymbol{\theta}_{rh}, \boldsymbol{\theta}_{lh}, \boldsymbol{\theta}_{b}\right]$, where $\boldsymbol{\theta}_{f} \in \mathbb{R}^{3+10}, \boldsymbol{\theta}_{rh},\boldsymbol{\theta}_{lh} \in \mathbb{R}^{15\times3}, \boldsymbol{\theta}_{b} \in \mathbb{R}^{21\times3}$ indicate pose parameters for face, right hand, left hand, and body, respectively.
To clarify, the target face parameters, jaw pose and expression, are combined in $\boldsymbol{\theta}_{f}$ for convenience.

{
The shape parameters are represented as $\boldsymbol{I} \in \mathbb{R}^{N_{s}}$ of the SMPL-X where $N_{s}$ is the shape dimension ($\leq 300$).
These encode the subject's physical identity such as bone length and body size.
}
In contrast to part-isolated input such as hand-hand~\cite{moon:eccv20} (MANO's) or hand-face interactions (MANO's + FLAME)~\cite{shimada:tog23}, the whole body parametrization provides additional constraints about the location of hands and face restricted through the kinematic chain of the body.
Owing to this, our representation only relies on local pose for simplicity while global orientation and translation of SMPL-X are disregarded.

\customparagraph{Latent diffusion with part-aware self-attention}
Regarding part-wise interaction modeling, we propose a part-aware self-attention transformer with latent embedding as $f$.
Unlike learning on pose parameters directly in the diffusion process~\cite{muller:cvpr24,lee:cvpr24}, we train the diffusion model in the latent space, inspired by latent diffusion for image synthesis~\cite{rombach:cvpr22}.
As joint movements in human motion are highly coordinated, the DOFs of whole-body joints can be represented in a lower-dimensional space~\cite{li:cvpr21,romero:tog17,li:cvpr23}.
Similarly, self-contact poses are intrinsically embedded in a latent manifold, as restricted to move along the body surface.
To enforce these constraints, we utilize auto-encoders for part-wise poses, enabling the discovery of plausible and semantically meaningful latent embeddings in training.

With the embeddings of the pose, diffusion time, and shape, we then utilize a self-attention transformer~\cite{vaswani:nips17} as the denoising module.
Specifically, the query, key, and value of the attention are given by the concatenation of the embeddings across face, right/left hand, body, time, and shape.
This enables the model to consider the interactions across part-wise poses, shape, and diffusion time (\ie, noisiness).
Part-wise learnable embeddings are also added to facilitate part-aware relational learning as~\cite{muller:cvpr24}.

We jointly train all the modules including the latent embedding and the transformer by following recent works~\cite{wu:3dv25,liu:icml24} that jointly train the parameters (or adapter parameters) of both networks to improve the
output quality.

\customparagraph{Shape-conditional perturbation}
Instead of naive conditioning of shape parameters, enriching the diversity of subjects in training is critical in learning the shape-dependent manifold and generalizing to unseen subjects in testing. 
Prior works use conditional dropout $\boldsymbol{c}=\varnothing$ to emulate unconditional generation~\cite{ho:arxiv22,lee:cvpr24,muller:cvpr24}, thereby increasing the diversity of generated samples.
However, in our context, a zero shape value corresponds to a plain body shape that lacks identity-specific signals.

We instead propose to perturb shape parameters slightly to augment the subject's identity, assuming that individuals with similar identities are likely to perform similar self-contact poses. 
As seen in ~\cref{fig:teaser}, two persons performing the ``rubbing belly'' pose differently, particularly the right arm angles are non-identical.
However, we observe that people with similar body shapes can come into contact with identical pose parameters.
We therefore replace a normal shape conditioning $\boldsymbol{c} = \boldsymbol{I}$ with the perturbed shapes with a certain probability (\eg, 30\%) as
\begin{equation}
    \boldsymbol{c} = \boldsymbol{I} + s_I \epsilon
\end{equation}
where $\epsilon$ is standard Gaussian noise and $s_{I}$ is a scaling factor to control the scale of the perturbation.

\subsection{Training objectives}\label{sec:loss}
The training loss is computed by taking the difference between the original data $\boldsymbol{X}_0$ and the generated data $\hat{\boldsymbol{X}}_0$.
We use the L1 loss between $\boldsymbol{X}_0$ and $\hat{\boldsymbol{X}}_0$ for the pose space, denoted as $\mathcal{L}_{\theta}$.
We also compute losses on the mesh space after constructing meshes with the SMPL-X layer.
We adopt the L1 loss between original and generated meshes for vertices $\mathcal{L}_{v}$, and an L1-based collision loss that penalizes vertices in collision on the generated mesh $\mathcal{L}_{col}$.
The overall loss is formulated as
\begin{equation}
    \mathcal{L}_{D} = \lambda_{\theta} L_{\theta} + \lambda_{v} L_{v} + \lambda_{col} L_{col}.
\end{equation} 
$\lambda_{\theta}$, $\lambda_{v}$, and $\lambda_{col}$ are the weights for each loss.
We use 6D rotation representation~\cite{zhou:cvpr19} for pose parameters.

\customparagraph{Collision detection}
The collision loss $\mathcal{L}_{col}$ is designed to avoid heavy penetration on the mesh, which is essential to maintain plausible self-contact.
For detection, \cite{muller:cvpr21} requires calculating pair-wise distances on vertices, but it is expensive to use on the fly in training.
This necessitates an efficient collision detector to consider collision in training.

We implement a fast ray-tracing-based collision detector following~\cite{smith:tog20}.
It casts rays in the normal direction from each vertex and computes the intersections with the mesh faces.
Counting the number of intersections serves to find the vertices inside the mesh.
Notably, vertices in the armpit region are often detected as collisions, leading to suboptimal solutions (\eg, forcing the arms to move far from the torso). To address this, we restrict collision detection to areas relevant to hands, which are explicit targets in self-contact.
Similarly to the data screening of \cref{sec:data}, we apply the loss only to penetrating hand vertices and their corresponding vertices, focusing on hand-specific collisions such as hand-over-hand, hand-over-belly, and hand-over-face.

\subsection{Inference}\label{sec:infer}
We describe data sampling in the following tasks. We use the DDIM sampling~\cite{song:iclr21} for efficiency.

\customparagraph{Random sampling}
The trained diffusion model allows random data sampling via the reverse process from random noise (T: $1000\rightarrow1$).
Inspired by~\cite{lee:cvpr24}, we reuse the collision loss in \cref{sec:loss} as the additional guidance term, \textit{anti-collision guidance}, which avoids the collision during the sampling phase as well. 
We set the sampling interval to 10.
This is used to produce pose generation results of \cref{sec:gen}.

\begin{algorithm}[t]
\caption{
\tbf{Single-view pose refinement}:
given initial 3D pose estimate $\boldsymbol{X}^{init}_0$, shape parameters $\boldsymbol{I}$,  2D keypoints to fit $\boldsymbol{P}_{2d}$, projection from pose to 2d keypoints $M_{p2d}$, a weight for 2d keypoint fitting $\lambda_{2d}$, start diffusion time $n_r$, mask for poses of interest $\boldsymbol{m}_p$.
}
\begin{algorithmic}[1]
\LeftComment{\textit{Initialization: diffuse 3D pose at step $n_r$}}
\State $\boldsymbol{X}_{n_{r}} \leftarrow \mathrm{noise}\left(\boldsymbol{X}^{init}_0, n_r\right)$
\For{$n={n_r}$ to 1}
        \State $\hat{\boldsymbol{X}}_0 \leftarrow f\left(\boldsymbol{X}_n, n, \boldsymbol{I}\right)$
        
        \Statex \hspace{0.5cm} \(\triangleright\) \textit{Optional: Blended pose denoising}
        \State $\hat{\boldsymbol{X}}_0 \leftarrow \hat{\boldsymbol{X}}_0 \odot \boldsymbol{m}_p + \boldsymbol{X}^{init}_0 \odot \left(1 - \boldsymbol{m}_p\right)$
        
        \State $\epsilon_n \leftarrow \frac{1}{\sqrt{1-\bar{\alpha}_n}}\left(\boldsymbol{X}_n-\sqrt{\bar{\alpha}_n} \hat{\boldsymbol{X}}_0\right)$
        \State $\boldsymbol{X}_{n-1}^{\prime} \leftarrow \sqrt{\bar{\alpha}_{n-1}} \hat{\boldsymbol{X}}_0+\sqrt{1-\bar{\alpha}_{n-1}} \epsilon_n$                
        \Statex \hspace{0.5cm} \(\triangleright\) \textit{2D keypoint fitting}
        \State $\boldsymbol{X}_{n-1} \leftarrow \boldsymbol{X}_{n-1}^{\prime}-\lambda_{2d} \nabla_{\boldsymbol{X}_n} \mathcal{L}_2\left(M_{p2d}(\hat{\boldsymbol{X}}_0), \boldsymbol{P}_{2d}\right)$        
        
\EndFor
\State \Return $\hat{\boldsymbol{X}}_0$
\end{algorithmic}\label{algo:svfit}
\end{algorithm}

\begin{table*}[t]
\centering
\renewcommand{\arraystretch}{1.1} %
\begin{minipage}{0.59\textwidth}
\centering
\resizebox{1.0\linewidth}{!}{
\begin{tabular}{c|cccccc}
\toprule
\textbf{Method}                    & \textbf{FID$\downarrow$} & \textbf{\begin{tabular}[c]{@{}c@{}}KID$\downarrow$ \\ ($\times10^{-3}$)\end{tabular}} & \textbf{Div.$\uparrow$} & \textbf{Prec.$\uparrow$} & \textbf{Recall$\uparrow$} & \textbf{\begin{tabular}[c]{@{}c@{}}Col. \\ratio$\downarrow$\end{tabular}} \\ \midrule
& \multicolumn{6}{|c}{{\textit{Unconditional generation}}}                                                \\                                                                                                                                                                                                                  
\multicolumn{1}{c|}{VPoser*~\cite{pavlakos:cvpr19}}        & 9.43                        & \underline{0.930}                                                                                     & 3.34                       & \underline{1.0}                        & 0.006                         & 1.72                                                                             \\
\multicolumn{1}{c|}{BUDDI*~\cite{muller:cvpr24}}        & \underline{3.19}                        & 1.16                                                                                     & \underline{6.36}                       & 0.957                        & \underline{0.528}                         & \underline{1.32}                                                                             \\ \hdashline

\vspace{-4mm} \\
& \multicolumn{6}{c}{\textit{Shape-conditional generation}}         \\     
\multicolumn{1}{c|}{VPoser*~\cite{pavlakos:cvpr19}}       & 9.16                        & 0.882                                                                                     & 3.20                       & \tbf{1.0}                        & 0.005                         & \tbf{1.37}                                                                             \\
\multicolumn{1}{c|}{BUDDI*~\cite{muller:cvpr24}}        & 2.66                        & 1.12                                                                                     & 5.59                       & 0.995                        & 0.488                         & 1.47                                                                             \\ \hdashline

\multicolumn{1}{c|}{\textbf{Ours}} & \tbf{1.25}                        & \tbf{0.430}                                                                                     & \tbf{5.98}                       & 0.985                         & \tbf{0.708}                                & 1.52                                                                  \\
\bottomrule
\end{tabular}
}
\caption{\textbf{Results of self-contact pose generation.}
We study sample quality and diversity in generation without (unconditional) or with shape conditioning, evaluated on the \textit{train} split.
The notation~* indicates the methods adapted to our task.
}
\label{tab:gen}
\end{minipage}
\hfill
\begin{minipage}{0.38\textwidth}
\centering
\small{
\begin{tabular}{l|ccc}
\toprule
\multicolumn{1}{c|}{\tbf{Method}}                                                        & \textbf{FID$\downarrow$} 
& \textbf{Div.$\uparrow$} 
& \textbf{\begin{tabular}[c]{@{}c@{}}Col. \\ratio$\downarrow$\end{tabular}} \\ \midrule
w/o Shape cond. & 2.18                       & 5.52                                   & 1.92                                                 \\\hdashline
w/o PASA                    & 1.42                        & 5.74                                                                                     & 1.62                                                                                             \\
w/o Shape rand.             & 1.27                        & 5.89                                                                                     & \tbf{1.41}                                                                                                 \\ 
w/o Anti-col.               & 1.28                        & \tbf{6.01}                                                                                     & 1.85                                                                                                \\\hdashline

\textbf{Ours}               & \tbf{1.25}                        & 5.98                                                                                    & 1.52                    \\         
\bottomrule
\end{tabular}
}
\caption{\textbf{Ablation study in our generation.}
We compare methods without shape conditioning (Shape cond.), part-aware self-attention (PASA) including a single latent space for pose, shape perturbation (Shape rand.), and anti-collision guidance (Anti-col.).
}
\label{tab:gen_ablation}
\end{minipage}
\vspace{-4mm}
\end{table*}

\customparagraph{Single-view pose refinement}
Observing self-contact poses inferred from single-view estimators~\cite{li:arxiv23,li:cvpr21,moon:cvprw22,cai:nips23}, 
the outputs often include incorrect contact states (\eg, hands not in touch) due to the lack of contact prior, while detected 2D keypoints are aligned well with the given image.
To address this, we develop single-view pose refinement, fitting the diffusion prior to the observed 2D keypoints with the estimated initial 3D poses.
This does not require additional training compared to score distillation sampling~\cite{poole:iclr23} of \cite{lee:cvpr24}, which is applicable to any 2D/3D estimates.

Our refinement is efficient in sampling with fewer sampling steps (\eg, only use the last 10\% steps); see \cref{algo:svfit}.
We assume given a single-view image, initial SMPL-X pose $\boldsymbol{X}^{init}_0$, shape $\boldsymbol{I}$, and 2D keypoints $\boldsymbol{P}_{2d}$ of the COCO-WholeBody format~\cite{jin:eccv20} can be estimated by off-the-shelf models, \eg, recent vision foundation models like SMPLer-X~\cite{cai:nips23} and Sapiens~\cite{khirodkar:eccv24}.
Then we sample data starting from the middle of the steps with diffused $\boldsymbol{X}^{init}_0$ at step $n_r$ (\eg, 100), reducing the number of sampling steps.
In each iteration, we use the guidance of 2D keypoint fitting to $\boldsymbol{P}_{2d}$ by computing the gradient of the 2D keypoint error (L2 loss).
We set the sampling interval to 1.
This is used to produce the refinement results of \cref{sec:sv}.

We also provide an additional option of \textit{blended pose denoising} in refinement.
We find the 2D observation is likely to be partially unavailable or unreliable from in-the-wild videos,
\eg, upper-body videos in video conferences do not provide 2D keypoint cues for the lower body.
Thus, inspired by image in-painting with diffusion models (\eg, Blended Latent Diffusion~\cite{avrahami:tog23}), we can only refine poses of interest (\eg, upper body poses) during the reverse process, while the rest of the poses are unchanged. 
Specifically, given the mask for blending $\boldsymbol{m}_p$, we replace poses not to be refined with $\boldsymbol{X}^{init}_0$ in each iteration (Line~4 of \cref{algo:svfit}), ensuring the convergence to the initial poses.
This simple trick helps control the inference process flexibly and enhances the applicability of the diffusion prior.

\section{Experiments}\label{sec:experiment}
We first present our dataset and implementation details in \cref{sec:exp-setup}, and then provide results for pose generation with random sampling and single-view pose estimation with our refinement method in \cref{sec:gen,sec:sv}.
We also show qualitative results in our proposed dataset and additional results in the supplement.

\subsection{Experiment setup}\label{sec:exp-setup}
\customparagraph{Datasets}
We create \train/\eval sets in the \DBname dataset.
The \train set is constructed with the captures with action instructions (\eg, rubbing neck), which comprises \DBtrain poses.
Additionally, the \eval set is designed for single-view pose estimation, featuring \textit{unseen subjects}, which contains \DBevalSubj samples.
This is used to test generalizability to unseen subjects where the same action instructions are given as the \train set. 

\customparagraph{Implementation details}
For generation, we use two self-attention layers with latent\_size=256, depth=4, num\_heads=4, and set $\lambda_{\theta}$=1, $\lambda_{v}$=1e-3, $\lambda_{col}$=1e-4.
{
We set $N_{s}$ to the full size of 300 to incorporate as much detailed shape information as possible.
}
We set the shape perturbation probability and $s_{I}$ to 0.3 and 1e-4.
Following~\cite{lee:cvpr24,raab:cvpr23,tevet:iclr23}, we report Fréchet Inception Distance (FID)~\cite{heusel:nips17}, Kernel Inception Distance (KID)~\cite{binkowski:iclr18}, diversity, and precision-recall~\cite{sajjadi:nips18} for the evaluation.
We also show the collision ratio of collided vertices over all SMPL-X vertices, using the detector of $\mathcal{L}_{col}$ in \cref{sec:loss}.

For single-view pose estimation, we prepare different SMPL-X regressors, namely HybrIK-X~\cite{li:arxiv23,li:cvpr21}, Hand4Whole~\cite{moon:cvprw22}, and SMPLer-X~\cite{cai:nips23}, and use Sapiens~\cite{khirodkar:eccv24} for 2D keypoint detection.
We report the MPJPE for 3D keypoints of the COCO-WholeBody format~\cite{jin:eccv20} in the body-root aligned coordinates (disregarding global rotation and translation).
We set $\lambda_{2d}$ and $n_r$ to 0.01 and 100.

\begin{figure*}[t]
\centering
\includegraphics[width=1\hsize]{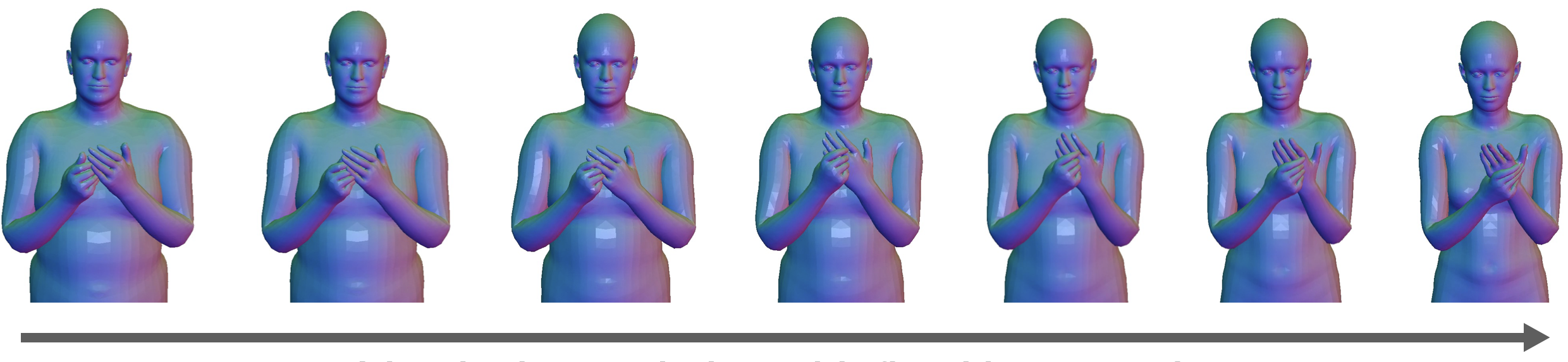}
\caption{\tbf{Qualitative results of our generation with shape interpolation}. We interpolate between two shape parameters 
with the fixed latent code (\ie, starting with the same noise at $t=T$).
Our model generates plausible self-contact poses under varying shapes. 
}
\vspace{-4mm}
\label{fig:shape_interp}
\end{figure*}

\subsection{Pose generation}\label{sec:gen}
\cref{tab:gen} shows pose generation results with or without shape conditioning in \DBname.
We compare our diffusion method with VAE-based VPoser~\cite{pavlakos:cvpr19} and diffusion-based BUDDI~\cite{muller:cvpr24}.
We modify these baselines to our task to take the whole body pose parameters as input (aligned to $\boldsymbol{X}$ of \cref{sec:model}), denoted as VPoser* and BUDDI*.
Details are found in our supplement.

The results show the superiority of our shape-conditional method, remarked by the improvement over the VAE and diffusion baselines.
While VPoser* easily overfits to higher precision, our diffusion method (Ours) has significantly improved recall with a smaller FID score.
Our method further surpasses BUDDI*, a method without latent diffusion and part-wise attention, achieving a 53\% reduction in FID.
Since random outputs exhibit higher diversity and samples lacking contact yield lower col.~ratio, maintaining higher diversity and lower col.~ratio scores with improved FID and KID (smaller distribution distance) is vital; our method shows a better balance between quality and variety.

\customparagraph{Ablation study}
\cref{tab:gen_ablation} shows the ablation study of our proposed method.
Shape-conditional generation surpasses unconditional baselines (w/o Shape cond.), which is also observed in the VPoser* and BUDDI* of \cref{tab:gen}.
This indicates that the body shapes help to learn self-contact pose distribution effectively.
The ablation study shows that part-aware self-attention (PASA), shape perturbation (Shape rand.), and anti-collision guidance (Anti-col.) reduce FID scores consistently over those without each module. 
Specifically, anti-collision guidance helps reduce col.~ratio in test time with improved FID. 

\customparagraph{Qualitative results}
\cref{fig:shape_interp} shows our qualitative results with shape interpolation. 
{
When changing the body shapes (from a large to a slim body), the generated poses continuously move on the hand surface while preserving plausible self-contact poses.
This indicates that our diffusion model can learn a smooth manifold of self-contact poses with respect to body shape changes.
}

\subsection{Single-view pose estimation and refinement}\label{sec:sv}
\customparagraph{Analysis on SMPL-X regressors}
\cref{tab:fit} shows single-view pose estimation results in \DBname.
We first evaluate existing SMPL-X regressors~\cite{li:arxiv23,moon:cvprw22,cai:nips23} in self-contact scenarios.
Hand4Whole and HybriIK-X adopt CNN-based backbones (\ie, ResNet~\cite{he:cvpr16} and HRNet~\cite{sun:cvpr19}). 
{
SMPLer-X~\cite{cai:nips23} is a foundation model trained on self-contact datasets (\ie, MTP~\cite{muller:cvpr21} and HumanSC3D~\cite{fieraru:aaai21}).
We further fine-tune the model on \DBname, denoted as SMPLer-X$^\dag$.
}

\begin{figure*}[t]
\centering
\includegraphics[width=1\hsize]{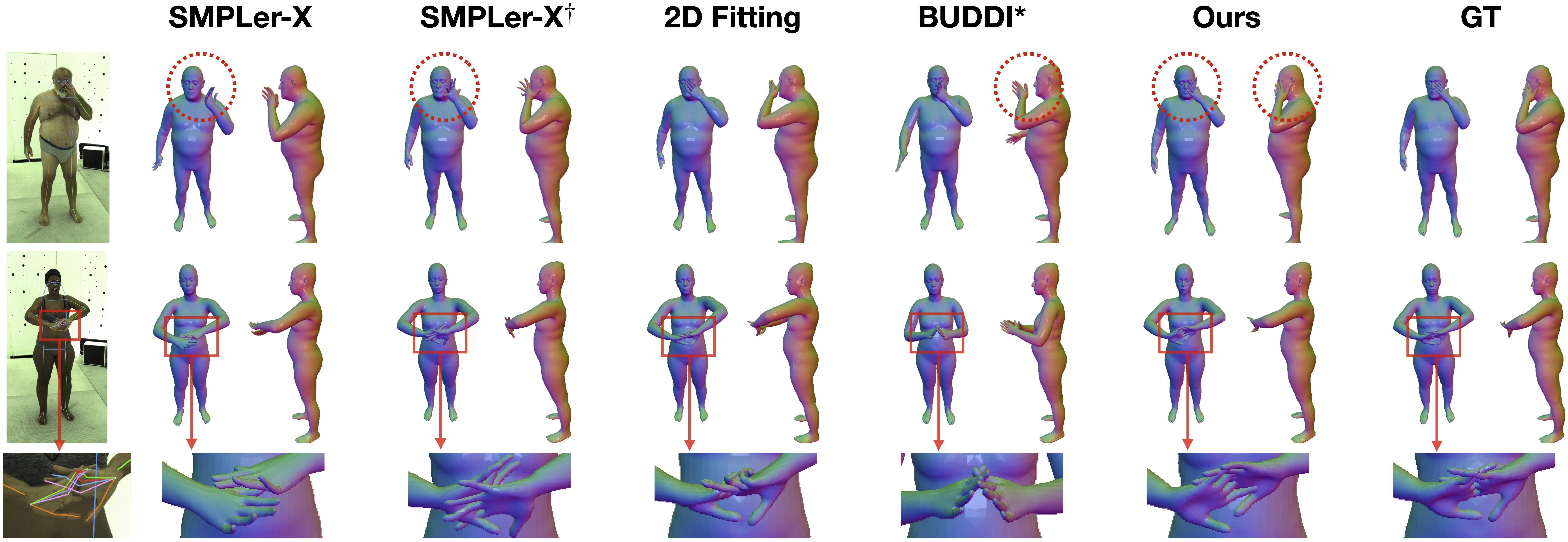}
\caption{\tbf{Qualitative results of our single-view refinement on \DBname.}
Our method successfully refines the initial poses to be valid self-contact for fine-grained poses, such as face touching and two-hand overlap.
}
\vspace{-4mm}
\label{fig:qual}
\end{figure*}

In the predictions from Hand4Whole and HybriIK-X, we find frequent failures in handling to place hands in contact, \ie, 2D pose is aligned in the image view but higher depth errors are present for hands.
Owing to the higher diversity in the training data, SMPLer-X facilitates tracking better poses in our \DBname data, with 58.0~mm error (see \cref{fig:qual}).
We confirm the state-of-the-art performance in image-based regression with the fine-tuned SMPLer-X$^\dag$, exhibiting an overall error reduction to 42.0~mm.

\begin{table}[t]
\renewcommand{\arraystretch}{1.1} %
\centering
\resizebox{1.0\linewidth}{!}{
\begin{tabular}{l|c:ccc}
\toprule
\tbf{Method}     & \tbf{Avg.} & \tbf{Hands} & \tbf{Body} & \tbf{Face} \\
\midrule
Hand4Whole~\cite{moon:cvprw22} & 126.3    & 225.8     & 89.6    & 78.0    \\ %
+ 2D fitting & 89.5 & 179.9  & 64.8 & 47.1 \\
+ BUDDI*~\cite{muller:cvpr24} & 74.5 & 109.2 & 37.8 & 65.9  \\\hdashline
+ \tbf{Ours (w/o Shape cond.)}        & 37.9    & 74.6     & 29.4    & \tbf{18.2}     \\ 
+ \tbf{Ours}                       & \tbf{35.3}    & \tbf{66.6}     & \tbf{26.5}    & 18.3   \\\midrule \midrule 

HybrIK-X~\cite{li:arxiv23}   & 82.3    & 99.2     & 62.8    & 76.2   \\ %
+ 2D fitting & 51.8 & 63.4  & 38.4  & 50.5\\
+ BUDDI*~\cite{muller:cvpr24} & 65.0 & 90.5 & 36.0  & 58.8 \\\hdashline
+ \tbf{Ours (w/o Shape cond.)}        & 45.9    & 85.5     & 32.5    & 26.3    \\
+ \tbf{Ours}                       & \tbf{32.4}    & \tbf{58.7}     & \tbf{26.1}    & \tbf{17.5}   \\\midrule 
\midrule
SMPLer-X~\cite{cai:nips23}   & 58.0    & 98.7     & 41.6    & 38.9    \\ %
SMPLer-X$^\dag$           & 42.0 & 56.7 & 31.9 & 34.1\\
+ 2D fitting        & 41.7 & 65.7 & 30.6 & 31.6 \\
+ BUDDI*~\cite{muller:cvpr24}  & 71.7    & 99.9     & 36.3    & 66.4  \\ \hdashline
+ \tbf{Ours (w/o Shape cond.)}        & 33.7    & 63.6     & 26.1    & \tbf{17.4}  \\
+ \tbf{Ours}                       &  \tbf{31.8}    & \tbf{54.6}     & \tbf{24.7}    & 19.2   \\

\bottomrule
\end{tabular}
}

\caption{\tbf{Results of single-view pose regression in \DBname.}
We evaluate our diffusion-based pose refinement in the \textit{eval} set
given initial pose estimation from SMPL-X regressors. We report MPJPE in millimeter on the body-root aligned coordinates.
The notation $^\dag$ shows fine-tuned results for the dataset.
}
\label{tab:fit}
\end{table}

\customparagraph{Analysis on refinement}
We compare our proposed refinement (\cref{algo:svfit}) with conventional refinement.
This setup assumes 2D/3D observations are given, namely initial SMPL-X estimates and 2D keypoints.
A naive baseline is a simple 2D keypoint fitting with optimization, which is 
widely used for pseudo-mesh registration of SMPL-X on in-the-wild videos~\cite{lin:cvpr23,moon:eccv24,hu:nips24}.
Since the body and face can be well-constrained with 2D keypoints, their gains are better than the fine-tuned results in the SMPLer-X setting.
In contrast, fitting to hands causes implausible 3D hand poses though the 2D projection error is minimized.
This underscores the need for the model prior in self-contact, particularly to correct hand placement and its local pose.

Next, we evaluate our refinement with diffusion-based priors, namely, our \Mname, and BUDDI* trained in \cref{tab:gen}.
BUDDI*, a method without latent diffusion and part-wise self-attention, shows effectiveness when the initial estimation is noisy (Hand4Whole and HybrIK-X), while it has limited refinement capacity when the initialization is reasonable (SMPLer-X).
This suggests that additional refinement of well-estimated poses requires more precise modeling of self-contact poses, which is essential to achieve improvements beyond the initial quality.

Our final diffusion prior achieves significantly improved results by reducing overall and part-wise errors across the three settings with different regressors.
Our refinement demonstrates less dependency on the initialization
and stable performance, as post-refinement converges to lower errors with the varied initializations.
Compared to the unconditional prior of our method (w/o Shape cond.), the proposed shape-conditional prior achieves better results, particularly for hands and body.
This indicates that the shape-dependent constraint is effective in capturing the relationship between the body and hands, as they have a higher correlation with body shapes than the face.

\customparagraph{Qualitative results}
\cref{fig:qual} shows qualitative results of our refinement.
We find that the initial predictions include ambiguities in contact and depth estimation, \eg, interacting parts are not in contact, especially for fine details, and high-depth errors remain for hands.
Our method can correct such failures with the generative prior derived from the contact data only, without requiring knowledge of where to contact.

\customparagraph{Discussion: regression \vs generative prior}
While recent 3D pose estimators are trained on extensive human data, the state-of-the-art baseline with regression still struggles to estimate self-contact poses of unseen subjects.
In contrast, our approach introduces a novel \textit{generative prior modeling} of self-contact pose distribution, with the body-shape dependent assumption.
Our model not only generalizes better to new subjects but also enhances robustness in handling fine-grained self-contact poses. 
This suggests that our generative prior offers a flexible and scalable solution for self-contact modeling over the regression approach.

\section{Conclusion}\label{sec:conclusion}
To highlight challenging self-contact scenarios, we offer a comprehensive self-contact analysis, along with the newly captured \tbf{\DBname} dataset with \DBpose poses and precise body shape registration.
We then model the self-contact pose manifold depending on body shapes with the generative diffusion model.
Specifically, the latent diffusion with part-aware self-attention learns pose distribution effectively and achieves the best results in pose generation.
We further propose single-view pose refinement using the diffusion prior, while fitting to the observed 2D keypoints.
Our experiments confirm the successful refinement of self-contact poses and show our superiority over the state-of-the-art diffusion method and the regressive foundation model.

\customparagraph{Limitation and future work}
{
We observe that hand-hand interaction is still a challenging scenario in the generation, in which minor interpenetration persists due to higher articulation, as studied in~\cite{lee:cvpr24}.
In addition, not only addressing in-contact scenarios only, but also generalizable modeling to non-contact cases like~\cite{muller:cvpr21} is an interesting extension.
}
Our success in self-contact modeling opens new avenues to include additional self-contact scenarios, \eg, without scripted action instructions, with various global body poses (\eg, sitting), or in multi-person conversation. 
Extending the diffusion prior into the temporal dimension or with linguistic contents is promising future work.

\customparagraph{Acknowledgment}
This work is partially supported by JST ASPIRE Grant Number JPMJAP2303 and JSPS KAKENHI Grant Numbers JP24K02956.

{\small
\bibliographystyle{cfgs/ieee_fullname}
\bibliography{main.bbl}
}

\section*{Appendix}

\section{Dataset Details}

\customparagraph{Data capture protocol}
Our dataset is constructed on the minimally-clothed body setup of~\cite{martinez:nips24}, which aligns with the previous work on body shape prior captured with subjects in tight-fit clothing~\cite{loper:tog15,pavlakos:cvpr19}.
We obtain user consent for data captures and the release of the registered SMPL-X parameters.

Due to the fixed capture space, most samples do not have high variations for lower-body poses.
Nevertheless, this setup allows for capturing \emph{intricate} upper-body self-contact details (\eg, ``rubbing eyes'') with unprecedented fidelity unavailable in existing studies~\cite{fieraru:aaai21,muller:cvpr21,fieraru:cvpr20,yin:cvpr23}.
While modeling large variations in lower-body pose (\eg, tying shoelaces) is not prioritized in this work, we will consider an expanded capture setup as future work.

\begin{figure}[t]
    \centering
    \includegraphics[width=1\linewidth]{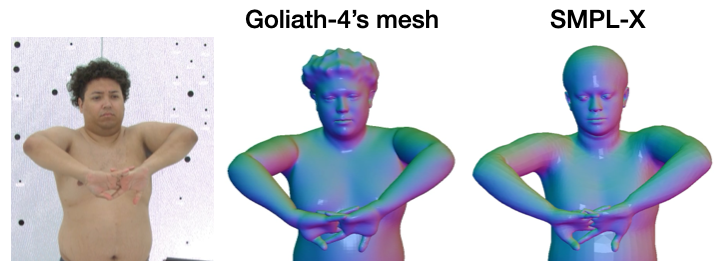}
    \caption{\tbf{Conversion from Goliath-4~\cite{martinez:nips24}'s mesh to SMPL-X.}}
    \label{fig:smplx_convert}
\end{figure}

\customparagraph{SMPL-X registration}
We initiate with a human mesh model used in ~\cite{martinez:nips24} that has a uniform topology across subjects, and we pre-compute its vertex-face correspondence to SMPL-X using barycentric coordinates. 
We register the human model across frames while tracking pose and surface precisely without relying on mocap markers.
Given multi-view dome captures, we first fit the human model to the rest pose (A-Pose).
Then we run 3D pose tracking based on multi-view images over the frames and use Linear Blend Skinning (LBS) that transforms a mesh in the rest pose to the desired pose of each frame.
The subject’s poses are continuously captured at 30~Hz with scripted action instructions to let participants express the corresponding gestures.
Given the registered mesh, the SMPL-X registration is obtained through vertex-to-vertex alignment between two meshes\footnote{
\url{https://github.com/vchoutas/smplx/blob/main/transfer_model/docs/transfer.md}
}, as shown in \cref{fig:smplx_convert}.
The continuous poses in our capture allow stable mesh alignment by using the previous frame's registration as initialization for the current frame, preventing significant fitting failures. 

\customparagraph{Data illustration}
Additional supplementary videos are included.
Videos~{\color{iccvblue}1-1} and~{\color{iccvblue}1-2} show our captured data, registered meshes, and contact maps from the Goliath-4's subjects~\cite{martinez:nips24}.
Video~{\color{iccvblue}2} illustrates more captured sequences. 
Video~{\color{iccvblue}3} is a video of the contact heatmap.
These include fine self-contact interactions with high-fidelity mesh registration.
The heatmap suggests a high contact likelihood on hands and across body parts, \eg, face, neck, belly, arm, back, and thigh.

\begin{figure*}[t]
\centering
    \includegraphics[width=0.9\hsize,page=1]{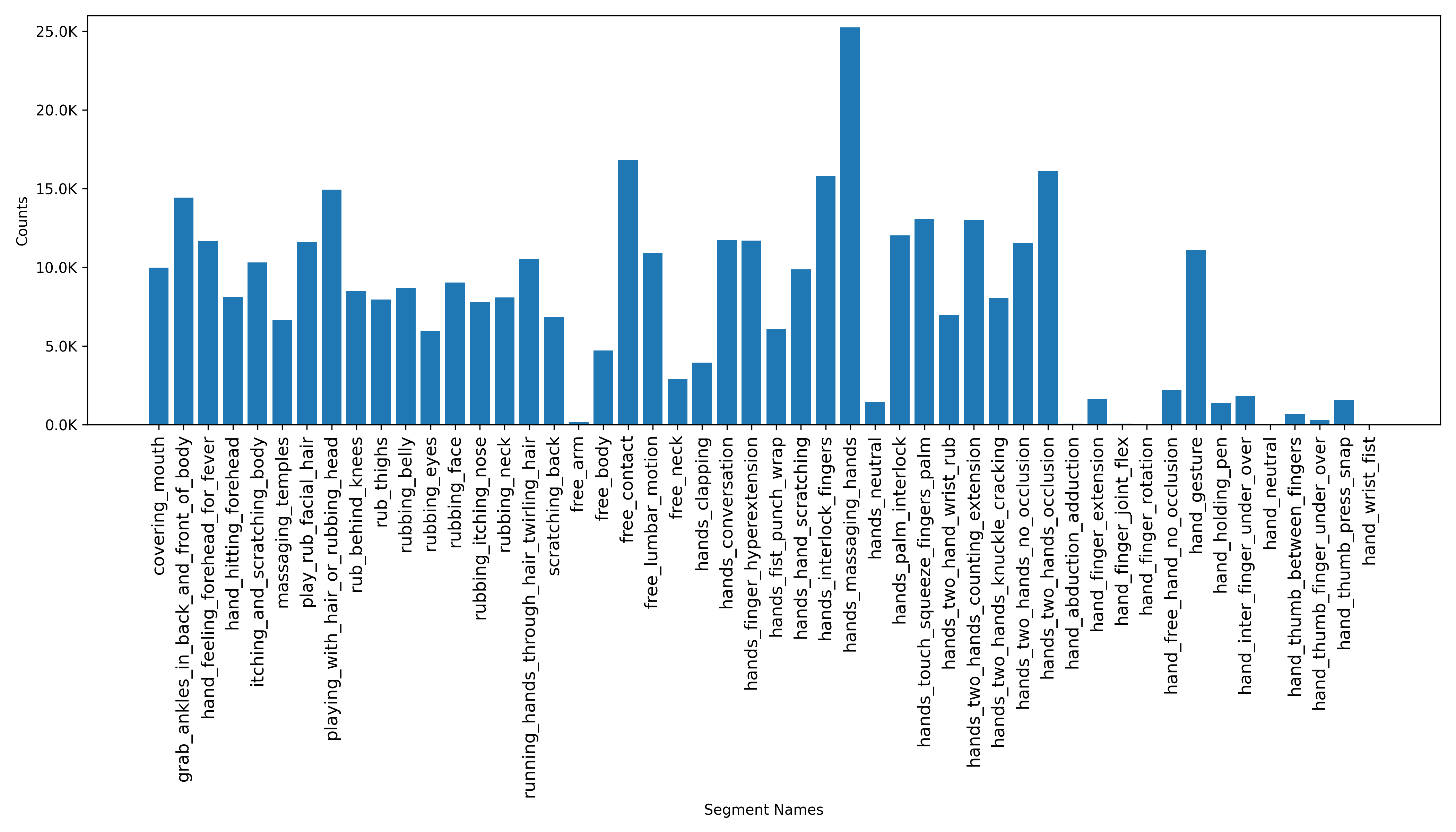}
\captionof{figure}{\tbf{Statistics of scripted actions and the number of self-contact poses in \DBname.}}
\label{fig:seg_stat}
\end{figure*}

\begin{figure*}[t]
    \centering
    \includegraphics[width=0.95\linewidth]{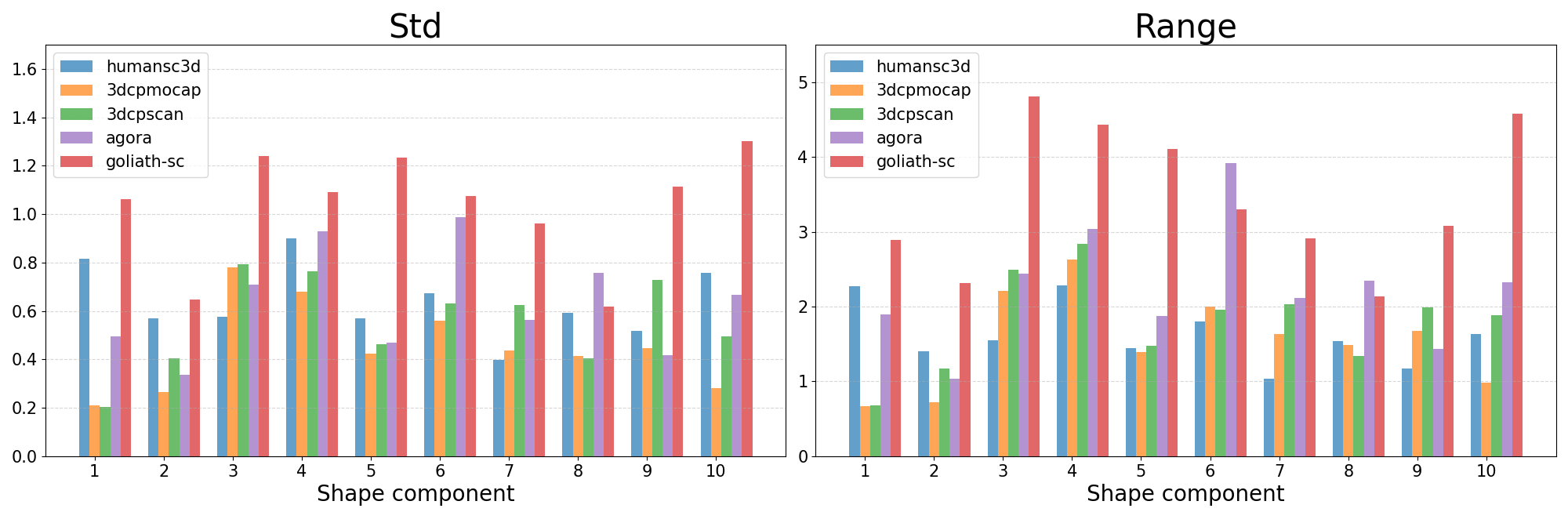}
    \caption{\tbf{Variability of subject shapes.}
    Standard deviation and range
    ($\text{max}-\text{min}$) 
    of the first 10 shape components in self-contact datasets, namely HumanSC3D~\cite{fieraru:aaai21}, (3DCPMocap, 3DCPScan, Agora) from MTP~\cite{muller:cvpr21}, and our Golaith-SC.
    }
    \label{fig:shape_dist}
\end{figure*}

\customparagraph{Data statistics}
\cref{fig:seg_stat} details action scripts used in the capture and the number of self-contact poses per action. 
The nouns of the actions suggest interacting body parts as following groups.
\begin{itemize}
    \item Head-related: Face, forehead, temples, eyes, nose, hair, facial hair, and neck
    \item Upper body-related: Arm, hand, wrist, fingers, thumb, and palm
    \item Torso-related: Belly, back, lumbar, and thighs
\end{itemize}
Instead, the verbs indicate how to interact with the body part; general movements are represented, such as hitting, grabbing, holding, clapping, rubbing, massaging, scratching, punching, wrapping, and itching.
In addition, hand-specific movements include extension, flex, rotation, press, snap, interlock, touch, and squeeze.
These hand interactions tend to be in close contact mostly in the captured sequence, resulting in a large number of poses in self-contact, such as ``hands massaging hands''.

Our dataset is constructed by capturing 130 subjects where the gender distribution is detailed in Tab.~{\color{iccvblue}1}.
To confirm the variety of captured shape information, we provide comprehensive analysis on shape statistics in ~\cref{fig:shape_dist}, \ie, standard deviation and range of 10 shape components of SMPL-X compared to the existing self-contact datasets, such as HumanSC3D~\cite{fieraru:aaai21} and MTP~\cite{muller:cvpr21}.
Our dataset (red) has the largest variety in most components except for 6th range, 8th std and range, indicating higher subject diversity and variability of our \DBname.

\section{Additional Implementation Details}
\customparagraph{Baselines}
We detail the implementation of baselines used in our experiments.
For fair comparison, we retrain the comparison models from scratch with the same input representation of the whole-body pose parameters (aligned to $\boldsymbol{X}$ of Sec.~{\color{iccvblue}4.2}), including hands, body, and face.

BUDDI~\cite{muller:cvpr24} is originally proposed for two-body interactions, modeling the joint distribution of the body pose parameters of SMPL-X (compatible to SMPL) and its shape parameters without latent diffusion modeling. 
A two-hand interaction generation model, InterHandGen~\cite{lee:cvpr24}, shares a similar architecture.
Our implemented BUDDI* modifies the original BUDDI to take the whole-body pose parameters of a single person. 
Following the original implementation, the transformer layers are used and all pose parameters are concatenated to a single vector, 
which indicates the absence of part-wise attention compared to our \Mname.
To produce the results of Tab.~{\color{iccvblue}2}, we construct baselines of the joint distribution modeling between pose and shape, \ie, unconditional model, and shape-conditional pose generation with the input embedding of the shape parameters.
In addition, when adapting to the task of single-view refinement, we use our fitting algorithm (Algorithm~{\color{iccvblue}1}) with the unconditional BUDDI* prior.
The hyper-parameters in the fitting (\eg, weight for 2D keypoint fitting and start diffusion time) follow those of our final method.

VPoser~\cite{pavlakos:cvpr19} is a VAE-based pose prior that learns pose distribution on the body pose parameters of SMPL-X.
Similarly to BUDDI*, we adapt this architecture for our task, by taking the whole-body pose parameters and adding shape parameters as conditional input, denoted as VPoser*.

\customparagraph{Training details}
We train generative models for 150,000 iterations with a batch size of 32, using an Adam optimizer~\cite{kingma:iclr14} with a learning rate of $\text{1e-4}$.
Our diffusion process is based on cosine noise scheduling with T=1000.
The auto-encoder network consists of a single linear layer with a hidden size of 256, which has separate weights for each body part.
Unlike the conventional choice of using 10 shape components of SMPL-X, we input the full 300-dimensional vector of the shape parameters to let the generative prior access as much fine details as possible (\eg, hands and face shapes).

\begin{figure*}[t]
\centering
\includegraphics[width=0.95\hsize]{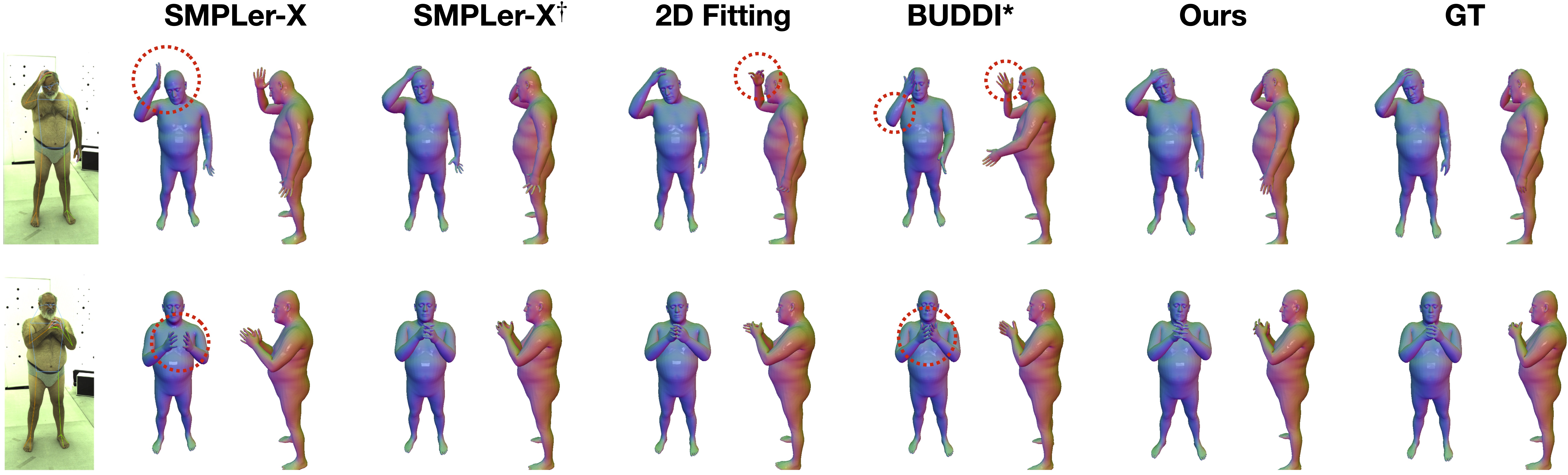}
\hfill
\includegraphics[width=0.95\hsize]{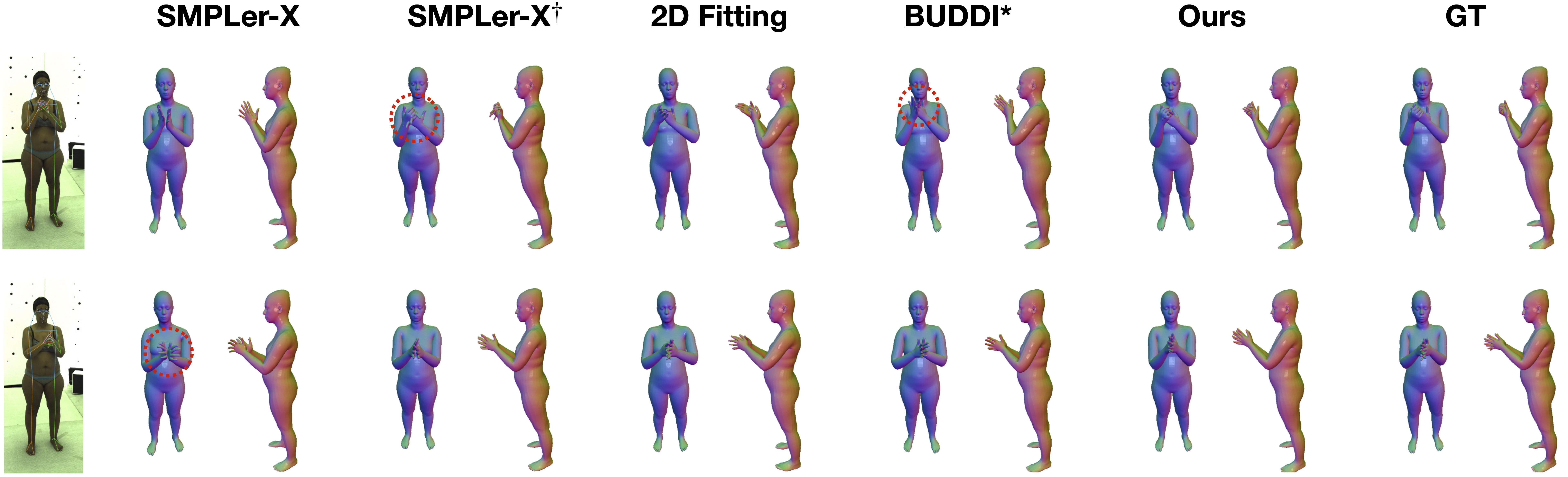}
\hfill
\includegraphics[width=0.95\hsize]{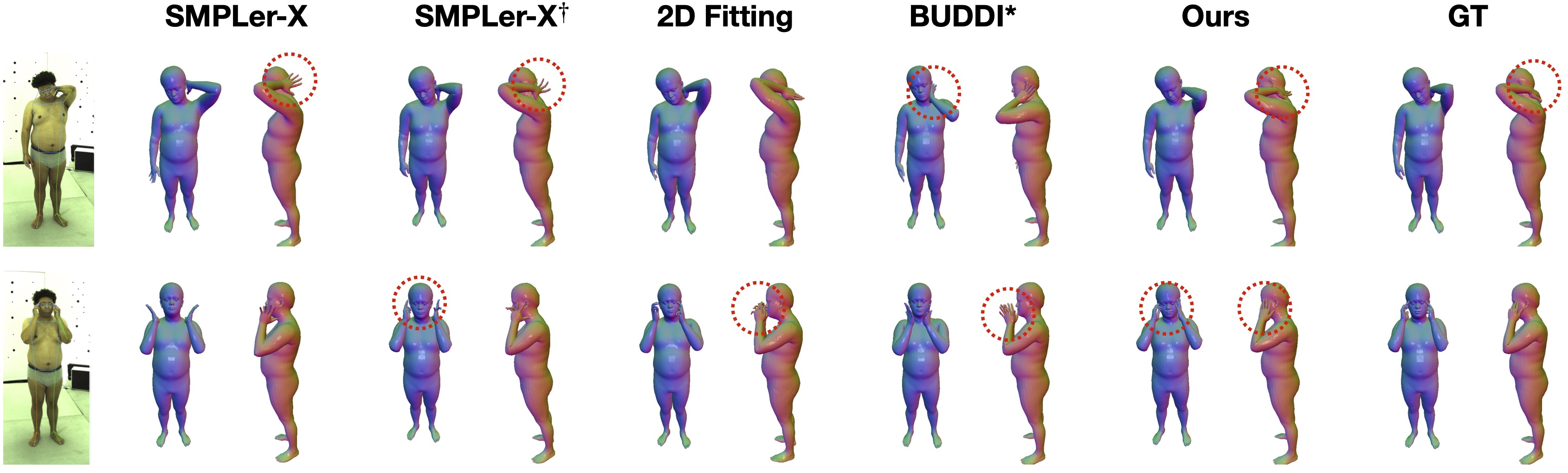}
\hfill
\includegraphics[width=0.95\hsize]{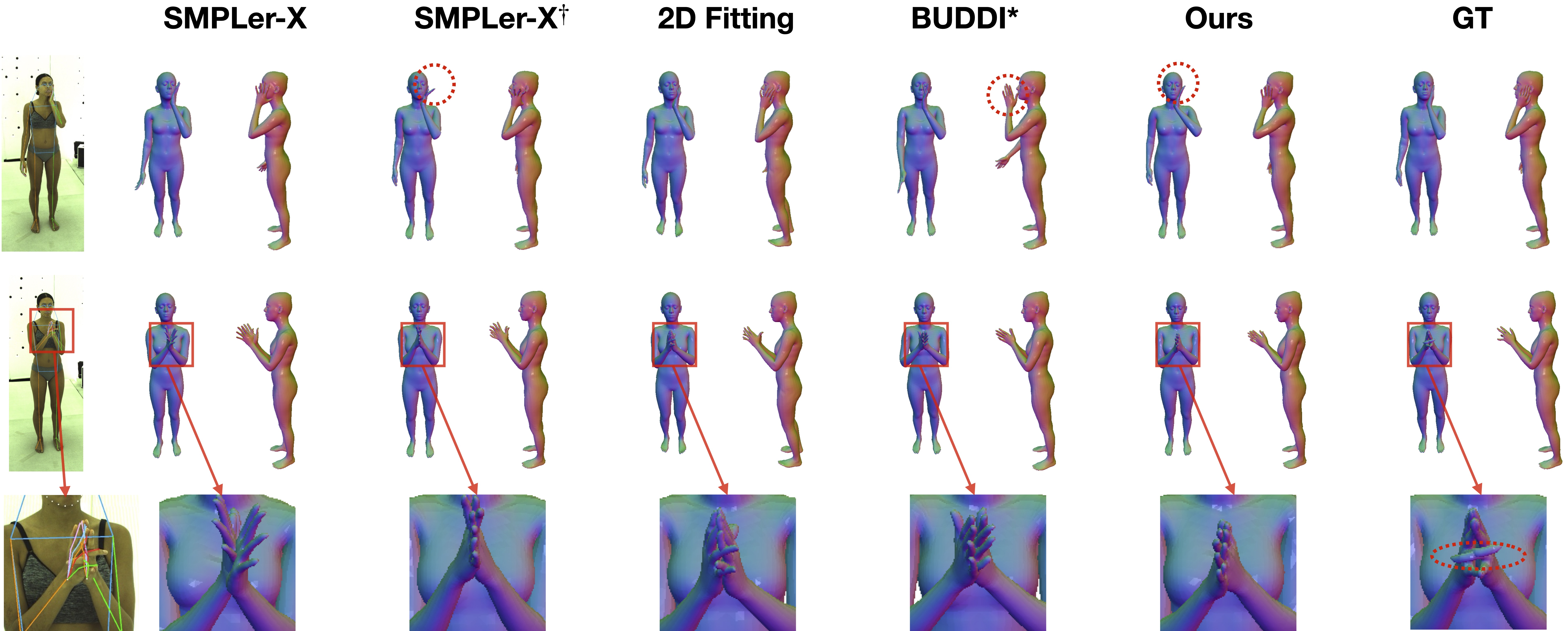}
\caption{\tbf{Qualitative results of single-view pose estimation.}
The four subjects of Goliath-4~\cite{martinez:nips24} are illustrated.
}
\label{fig:sv_fit4}
\end{figure*}

\section{Additional Results}

\customparagraph{Qualitative results of shape interpolation}
Video~{\color{iccvblue}4} shows additional results of shape interpolation with our shape-conditional \Mname.
Poses are sampled from interpolated shape parameters and fixed noise input, and four self-contact videos are concatenated into the Video~{\color{iccvblue}4}.
The results indicate that different noise inputs correspond to non-identical self-contact patterns. 
With the shape changes, we find that the interacting parts are almost consistent and no significant corruption is observed.
This suggests that the proposed diffusion prior enables learning a smooth pose manifold dependent on the given shapes.

\customparagraph{Qualitative results of single-view pose estimation}
Additional qualitative results of single-view pose estimation are found in~\cref{fig:sv_fit4}, including SMPLer-X, fine-tuned SMPLer-X$^\dag$, 2D fitting, BUDDI*, our final refinement (Ours), and GTs.
We observe that hands are not often in contact with SMPLer-X (\eg, Rows~1,2,4,6), while the fine-tuned baseline struggles with highly bent hand fingers (\eg, Rows~3,6,7) and incorrect contact states, \eg, for the hidden left hand behind the neck of Row~5. 
The 2D keypoint fitting baseline tends to exhibit unsolved depth ambiguity (Rows~1,6) and implausible hand poses (Row~1) due to overfitting to the 2D observation. 
The BUDDI* method often relies heavily on the model prior with a large 2D error to the observation.
This indicates that the method generates plausible poses, yet not aligned to the given 2D keypoints, such as Rows~1,2,3,5.
It also comprises higher hand depth errors (to those to be in contact) like Rows~1,2,6.
Notably, our method can resolve these failures presented by the comparison models and shows significantly reduced errors in 3D compared to the GT.

The last row shows a remaining failure when fingers are in complex interaction, \ie, the fingers of both hands, except for the ring fingers, are overlapped while only the ring fingers are bent.
Neither method handles this pose well because of the inaccuracy of 2D keypoint detection.
Improving detection and model-based refinement to such fine interactions are future challenges.

\end{document}